\documentclass[letterpaper, 10 pt, conference]{ieeeconf}  

\IEEEoverridecommandlockouts                              

\usepackage{graphics} 
\usepackage{graphicx,subfigure}
\usepackage{epsfig} 
\usepackage{amsmath} 
\usepackage{amssymb}  
\usepackage{multirow} 
\usepackage{algorithm}
\usepackage{algorithmic}
\usepackage{stfloats}

\title{\LARGE \bf
Towards cognitive exploration through deep reinforcement learning \\
for mobile robots
}

\author{Lei Tai$^{1}$ and Ming Liu$^{1,2}$
\thanks{$^{*}$This work was supported by the Research Grant Council of Hong Kong SAR Government, China, under project No. 16206014 and No. 16212815; National Natural Science Foundation of China No. 6140021318, awarded to Prof. Ming Liu}
\thanks{$^{1}$Lei Tai and Ming Liu are with MBE, City University of Hong Kong.
        {\tt\small lei.tai@my.cityu.edu.hk}}%
\thanks{$^{2}$Ming Liu is with  ECE, Hong Kong University of Science and Technology.
        {\tt\small eelium@ust.hk}}%
}

\begin{document}

\maketitle
\thispagestyle{empty}
\pagestyle{empty}

\begin{abstract}

Exploration in an unknown environment is the core functionality for mobile robots. Learning-based exploration methods, including convolutional neural networks, provide excellent strategies without human-designed logic for the feature extraction \cite{tai2016deep}. But the conventional supervised learning algorithms cost lots of efforts on the labeling work of datasets inevitably. Scenes not included in the training set are mostly unrecognized either. 
We propose a deep reinforcement learning method for the exploration of mobile robots in an indoor environment with the depth information from an RGB-D sensor only. Based on the Deep Q-Network framework \cite{mnih2015human}, the raw depth image is taken as the only input to estimate the Q values corresponding to all moving commands. The training of the network weights is end-to-end. In arbitrarily constructed simulation environments, we show that the robot can be quickly adapted to unfamiliar scenes without any man-made labeling. Besides, through analysis of receptive fields of feature representations, deep reinforcement learning motivates the convolutional networks to estimate the traversability of the scenes. The test results are compared with the exploration strategies separately based on deep learning \cite{tai2016deep} or reinforcement learning \cite{tl_rcar_2016}. Even trained only in the simulated environment, experimental results in real-world environment demonstrate that the cognitive ability of robot controller is dramatically improved compared with the supervised method. We believe it is the first time that raw sensor information is used to build cognitive exploration strategy for mobile robots through end-to-end deep reinforcement learning.

\end{abstract}

\section{Introduction}

\subsection{Motivation}
\label{sec:motivation}
Exploration in an unknown environment is fundamental for mobile robots for tasks like cleaning, mining, and rescue, etc. By this work, we deal with a classic task for a mobile robot equipped with a depth sensor: attempt to explore as much area as possible in an unknown environment while avoiding collisions with obstacles. Conventional exploration methods require heuristic control logic such as 
the front-wave exploration \cite{arkin1990integrating} and additional processes to deal with obstacles \cite{borenstein1991vector}. Aided by stereo vision systems or radar sensors, researchers often build the geometry or topological mapping of environments \cite{liu2015incremental} \cite{liu2012markov} to make navigation decisions based on a global representation. These methods look at the environment as a geometrical world and decisions are only made with preliminary features without a cognitive process. Specific logic has to be particularly designed for different environments. It is still a challenge to rapidly adapt to a new environment for mobile robots.

Convolutional neural networks (CNN), a typical model for deep-learning and cognitive recognition, have taken the state-of-the-art place in computer vision tasks. Successes of this hierarchical model also motivate robotic scientists to apply deep learning algorithms in conventional robotics problems like recognition \cite{hou2015convolutional} and obstacle avoidance \cite{tai2016deep} \cite{giusti2016machine} \cite{muller2005off}.

As the same as most of the supervised learning algorithms, CNN extracts feature representations through training with a huge amount of labeled samples. Nevertheless, unlike typical computer vision tasks, robotics exploration usually happens in dynamical environments with higher probability and uncertainty. The overfitting problem of supervised learning limits the perception ability of hierarchical models for the untrained inputs, and it is unrealistic for mobile robots to collect datasets covering all of the possible conditions. Besides, the time-consuming work of datasets collection and labeling seriously influences the application of CNN-based learning methods. Another commonly recognizable problem is that the robotic research usually considers the mechanism of CNN as a black box. It lacks a proper metric to validate the efficiency of the network and let alone the improvement. We will use the receptive fields to visually show the salient regions that determine the output. It provides the ground for structure selection and performance justification. 

Reinforcement learning is such an efficient way to learn control policies without referencing the ground-truth. Through combining reinforcement learning and hierarchical sensory processing, deep reinforcement learning (DRL) \cite{mnih2015human} can learn optimal policies directly from high-dimensional sensory inputs. And it outperformed all of the previous artificial control algorithms in Atari games \cite{mnih2013playing}. 

In our previous work, we have proved the feasibility of the CNN-based supervised learning method for obstacle avoidance in the indoor environment \cite{tai2016deep} and the effectiveness of the conventional reinforcement learning method in the exploration policy estimation \cite{tl_rcar_2016} through the feature representations extracted from the pre-trained CNN model in \cite{tai2016deep}. In this paper, we propose an end-to-end deep reinforcement learning method towards cognitive exploration in an unfamiliar environment by taking depth image as the input and control commands as the output. Not like conventional learning methods, the training of deep reinforcement learning is a cognitive process. The optimization of this exploration policy is incremental with the training going for mobile robots.

\subsection{Contributions}
\label{sec:contributions}
We stress the following contributions and features of this work:


\begin{itemize}

\item 
By deep reinforcement learning, we show the developed exploration capability of a mobile robot in unknown environments. We initialize the weights from the previous CNN model trained with real-world sensory samples and continually train it in an end-to-end manner. The performance is evaluated in both simulated and real-world environments.

\item The deep reinforcement learning model can quickly achieve obstacle avoidance ability in an indoor environment with several thousands of training iterations, without additional man-made collection or labeling work for datasets. 

\item For evaluations of CNN, we use receptive fields in origin inputs to reason the feasibility of the trained model. The receptive fields activated by the final feature representations are presented through bilinear upsampling. The activation characters prove the cognitive ability improvement of hierarchical convolutional structures for 
traversability estimation.

\end{itemize}


\section{Related Work} \label{sec:rel}

Conventional robot exploration strategies mainly depended on complicated control logics, with hand-crafted features extracted from environments \cite{kuipers1991robot}. Benefiting from the development of large-scale computing hardware like GPU, deep learning related methods have been considered to address robotics related problems including robot exploration.

\subsection{Deep learning in robotics exploration}
Convolutional neural networks (CNNs) have been applied to recognize off-road obstacles \cite{muller2005off} by taking stereo images as input. It also helped aerial robotics to navigate along forest trails with a single monocular camera \cite{giusti2016machine}. In our previous work, a three-layer convolutional framework \cite{tai2016deep} was used to perceive an indoor corridor environment for mobile robots. By taking raw images or depth images as inputs, and taking the moving commands or steering angles as outputs, weights of CNN-based models can be trained through back-propagation and stochastic gradient descent. Except for robotics exploration, grasping locations can be regarded as an object detection problem \cite{lenz2015deep} and CNN is the state-of-the-art solution for this problem.

Notice that all of these supervised learning methods mentioned above require a large amount of efforts on collecting and labeling of datasets. Kim \textit{et al}. \cite{kim2006traversability} achieved the labeling result by using other sensors with higher resolution. Tao \textit{et al}. \cite{tao2015semi} labeled the center sample of the clustering result for object classification as a semi-supervised method. Considering the requirement for auxiliary judgments, unsupervised learning methods didn't eliminate the labeling work essentially.  

The huge potential of deep learning in raw image processing has shown great probability to solve visual-based robot control problems. 
However, even though CNN related methods accomplished lots of breakthroughs and challenging benchmarks for vision perception tasks like object detection and image recognition, applications in robotics control are still less prevalent.

\subsection{Reinforcement learning in robotics}

Reinforcement learning is a useful way for robotics to learn control policies. The main advantage of reinforcement learning is the completed independence from human-labeling. Motivated by the trial-and-error interaction with the environment, the estimation of the action-value function is self-driven by taking the robot states as the input of the model. Conventional reinforcement learning methods improved the controller performances in path-planning of robot-arms \cite{xie2015model} and controlling of helicopters \cite{ng2006autonomous}. 

Through regarding RGB or RGB-D images as the states of robots, reinforcement learning can be directly used to achieve visual control policies. In our previous work \cite{tl_rcar_2016}, a Q-learning based reinforcement learning controller was used to help a \textit{turtlebot} navigate in the simulation environment.  

\subsection{Deep reinforcement learning}

Due to the potential of automating the design of data representations, deep reinforcement learning abstracted considerable attentions recently \cite{duan2016benchmarking}. Deep reinforcement learning was firstly applied on playing 2600 Atari games \cite{mnih2015human}. The typical model-free Q-learning method was combined with convolutional feature extraction structures as Deep Q-network (DQN). The learned policies beat human players and previous algorithms in most of Atari games. Based on the success of DQN \cite{mnih2015human}, revised deep reinforcement learning methods appeared to improve the performance on various of applications. Not like DQN taking three continues images as input, DRQN \cite{hausknecht2015deep} replaced several normal convolutional layers with recurrent neural networks (RNN) and long short term memory (LSTM) layers. Taking only one frame as the input, the trained model performed as well as DQN in Atari games. Dueling network \cite{wang2015dueling} separated the Q-value estimator to two independent network structures, one for the state value function and one for the advantage function. Now, it is the state-of-art method on the Atari 2600 domain. 

For robotics control, deep reinforcement learning also accomplished various simulated robotics control tasks \cite{lillicrap2015continuous}. In the continues control domain \cite{gu2016continuous}, the same model-free algorithm robustly solved more than 20 simulated physics tasks. Control policies are learned directly from raw pixel inputs. Considering the complexity of control problems, model-based reinforcement learning algorithm was proved to be able to accelerate the learning procedure \cite{lillicrap2015continuous} so that the deep reinforcement learning framework could handle more challenging problems. 

No matter Atari games or the control tasks mentioned above, deep reinforcement learning has been keeping showing the advantage in simulated environments. However, it is rarely used to address robotics problems in real world environment. As in \cite{zhang2015towards}, the motion control of a \textit{Baxter} robot motivated by deep reinforcement learning could make sense only with simulated semantic images but not raw images taken by real cameras. Thus, we consider the feasibility of deep reinforcement learning in real world tasks to be the primary contribution of our work.

\section{Implementation of deep reinforcement learning} \label{sec:drl}
One of the main limitations of applying deep reinforcement learning in real world environment is that the repeated actor-critic learning procedure of reinforcement learning is quite difficult to be implemented in the actual world. As in Atari games, the controller must repeat the games thousands of times after each attempt episode. But in a real physical world, it is unrealistic for robotics to repeat the same task with the same beginning state again and again. In this paper, we implement the same model-free deep reinforcement learning framework like DQN \cite{mnih2015human} in the simulation environment as well. To help the convergence, the weights of convolutional neural networks are initialized from a supervised learning model with data collected from the actual environments. In the end-to-end training procedure, we set a small learning rate for the gradient descent of data representation structure compared with the learning rate used in the training of the supervised learning model \cite{tai2016deep}. Finally, the learned model can both keep the navigation ability in the origin world and build the adaptation for an unknown world.

\subsection{Simulated environment} \label{sec:simuenvi}

\begin{figure}[!h]
    \centering
   \includegraphics[width=0.8\columnwidth]{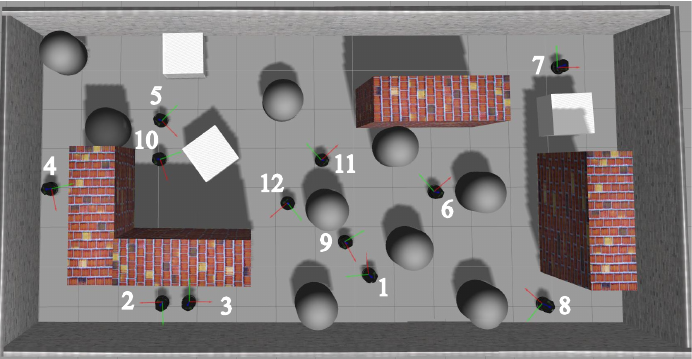}
    \caption{The simulated indoor environment implemented in \textit{Gazebo} with various scenes and obstacles. The \textit{turtlebot} is the experimental agent with a \textit{kinect} camera mounted on it. One of the 12 locations marked in the figure is randomly set as the start point in each  training episode. The red arrow of every location represents the initial moving direction.}   
   \label{fig:environment_figure}
\end{figure}

In our previous work \cite{tai2016deep}, the training datasets were collected in structured corridor environments. Depth images in these datasets were labeled with real-time moving commands from human decisions. In this work, to extend the exploration ability of the mobile robot, we set up a more complicated indoor environment as shown in Fig.~\ref{fig:environment_figure} in the \textit{Gazebo}\footnote{http://gazebosim.org/} simulator. Besides the corridor-like traversable areas, there are much more complicated scenes like cylinders, sharp edges and multiple obstacles with different perceptive depths. These newly created scenes have never been used in the training of our previous supervised learning model \cite{tai2016deep}.

We use a \textit{turtlebot} as the main agent in the simulated environment. A \textit{kinect} RGBD camera is mounted on top of the robot. We can receive the real-time RGB-D raw image from the field of view (FOV) of the robot. All of the requested information and communications between agents are achieved through \textit{ROS} \footnote{http://www.ros.org} interfaces.

\subsection{Deep reinforcement learning implementation}
   \begin{figure}[!ht]
      \centering
      \includegraphics[width=0.75\columnwidth]{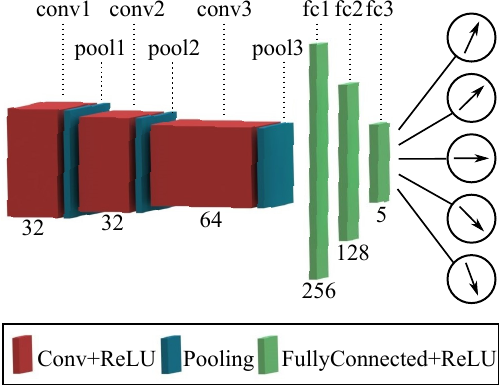}
      \caption{The network structure for the actor-evaluation estimation. It's a combination of the convolutional networks for feature extraction and the fullyconnected layers for policy learning. They have been separately proven to be effective in our previous work \cite{tai2016deep} \cite{tl_rcar_2016}.}
      \label{fig:network_structure}
   \end{figure}

As a standard reinforcement learning structure, we set the environment mentioned in Section \ref{sec:simuenvi} as $e$. At each discrete time step, the agent selects an action $ a_t $ from the defined action set. In this paper, the action set consists of five moving commands, namely \textit{left, half-left, straight, half-right} and \textit{right}. Detailed assignments of speeds related to the moving commands are introduced in Section \ref{sec:experiment}. The only perception by the robot is the depth image $x_t$ taken from the \textit{kinect} camera after every execution of the action. Unlike that the reward $r_t$ in Atari games is the change of the game's score, the only feedback used as the reward is a binary state, indicating whether the collision occurs or not. It is decided by checking the minimum distance $l_t$ through the depth image taken by the \textit{kinect} camera. Once the collision occurs, we set a negative reward $t_{ter}$ to represent the termination. Conversely, we grant a positive reward $t_{move}$ to encourage the collision-free movement.

The exploration sequences $s_t$ in the simulated environment is regarded as a \textit{Markov Decision Process} (MDP). It is an alternate combination of moving commands and depth-image states where $s_t = \{x_1,a_1,x_2,a_2,\dots, a_{t-1},x_t\}$. The sequence terminates once the collision happens. As the assumption of MDP, $x_{t+1}$ is completely decided by $(x_t, a_t)$ without any references with the former states or actions in $s_t$. The sum of the future rewards until the termination is $R_t$. With a discounted factor $\gamma$ for future rewards, the sum of future estimated rewards is $R_t = \sum_{t'=t}^{T} \gamma^{t'-t} r_{t'}$, where $T$ means the termination time-step. The target of reinforcement learning is to find the optimal strategy $\pi$ for the action decision through maximizing the action-value function $Q^*(x,a) = max_{\pi}\mathbb{E}[R_t|x_t=x,a_t=a,\pi]$. The essential assumption in DQN \cite{mnih2015human} is the \textit{Bellman equation}, which transfers the target to maximize the value of $r+\gamma Q^{*}(x',a')$ as
\[ Q^{*}(x,a)=\mathbb{E}_{{x'}\sim{e}}[r+\gamma \max \limits_{a'}Q^{*}(x',a')|x,a] \]
Here $x'$ is the state after acting action $a$ in state $x$.
DQN estimated the action-value equation by convolutional neural networks with weights $\theta$, so that $Q(s,a,\theta) \approx Q^{*}(s,a)$. 
\begin{algorithm}[!ht]
    \caption{Deep reinforcement learning algorithm}
    \label{alg:training_al}
    \begin{algorithmic}[1]
    \STATE {Initialize the weights of evaluation networks as $\theta^-$\\
           Initialize the memory $D$ to store experience replay\\
           Set the collision distance threshold $l_s$
           }
    \FOR {episode $ = 1, M$}
    \STATE {Randomly set the \textit{turtlebot} to a start position\\
            Get the minimum intensity of depth image as $l_t$\\
        \WHILE{$l_t>l_s$}
            \STATE { Capture the depth image $x_t$ }
            \STATE {With probability $\varepsilon$ select a random action $a_t$ \\
                    Otherwise select $ a_t =\mathrm{argmax}_a Q(x_t,a;\theta^-)$ \\}
            \STATE { Move with the selected moving command $a_t$\\
                    Update $l_{t}$ with new depth information\\
            \IF{$l_{t}<l_s$}
                \STATE{ $r_t =r_{ter}$ \\
                $x_{t+1} = Null$}
            \ELSE
                \STATE{ $ r_t = r_{move}$ \\
                Capture the new depth image $ x_{t+1}$}
            \ENDIF \\
            }
            \STATE{
            Store the transition $(x_t,a_t,r_t,x_{t+1})$ in $D$} \\            
            Select a batch of transitions  $(x_k, a_k, r_k, x_{k+1})$  randomly from $D$ \\
                \IF {$r_k = r_{ter}  $}
                    \STATE{ $y_k = r_k$   }
                \ELSE
                \STATE{  $ y_k = r_k+\gamma \max_{a'} Q(x_{k+1},a';\theta^-) $ }
                \ENDIF \\
           Update $\theta$ through a gradient descent procedure on the batch of $ (y_k - Q(\phi_k,a_k;\theta^-))^2$
        \ENDWHILE}
    \ENDFOR
\end{algorithmic}
\end{algorithm}

In this paper, we use three convolutional layers for feature extractions of the depth image and use additional three fully-connected layers for exploration policy learning. The structure is depicted as red and green cubes shown in Fig.~\ref{fig:network_structure}. To increase the non-linearity for better data fitting, each Conv or Fully-connected layer is followed by a Rectified Linear Unit (ReLU) activation function layer.
The number under each Conv+ReLU or FullyConnected+ReLU cube is the number of channels of the output data related to this cube. The network takes a single depth raw image as the input. The five channels of the final fully-connected layer \textit{fc3} are the values of the five moving commands. Besides, to avoid the overfitting in the training procedure, both of the first two fully-connected layers \textit{fc1} and \textit{fc2} are followed with a dropout layer. Note that dropout layers are eliminated in test procedure \cite{srivastava2014dropout}.

Algorithm \ref{alg:training_al} shows the workflow of our revised deep reinforcement learning process. Similar as \cite{mnih2015human}, we use the memory replay method and the $\varepsilon$-greedy training strategy to control the dynamic distribution of training samples. After the initialization of the weights for convolutional networks shown in Fig.~\ref{fig:network_structure}, set a distance threshold $l_s$ to check if the \textit{turtlebot} collides with any obstacles. At the beginning of every repeated exploration loop, the \textit{turtlebot} is randomly set to a start point among the 12 pre-defined start points shown in Fig.~\ref{fig:environment_figure}. That extends the randomization of the \textit{turtlebot} locations from the whole simulation world and keeps the diversity of the data distribution saved in memory replay for training.

For the update of weights $\theta$, $y_k$ is the target for the evaluation network to output. It is calculated by summing the instant reward and the future expectation estimated by the networks with the former weights as mentioned before in the \textit{Bellman equation}. If the sampled transition is a collision sample, the evaluation for this $(x_k , a_k)$ pair is directly set as the termination reward $r_{ter}$. Setting the training batch size to be $n$, the loss function is 
\[    
L({\theta}_i) = \frac{1}{n} \sum_{k}^{n}[(y_k-Q(x_k,a_k;{\theta}_i))^2]
\]
After the estimation of $Q(x_k, a_k)$ and $\max_{a'} Q(x_{k+1},a')$ with the former $\theta^-$, the weights $\theta$ of the network will be updated through back-propagation and stochastic gradient descent.

\section{Experiments And Results} \label{sec:experiment}
\subsection{Training}
At the beginning of the training, Convolutional layers are initialized by copying the weights trained in \cite{tai2016deep} for the same layer structure. A simple policy learning networks structure was also separately proved in \cite{tl_rcar_2016} with three moving commands as output. 
\begin{table}[!ht]
    \centering
    \caption{Training parameters and the related values}
    \label{tab:training_parameters}
    \begin{tabular}{l r}
    \hline
    \hline
    Parameter  & Value\\
    \hline
    batch size&32\\
    replay memory size & 3000\\
    discount factor $\gamma$ & 0.85\\
    learning rate & 0.0000001\\
    gradient momentum & 0.99\\
    distance threshold  $l_s$ & 0.55\\
    negative reward  $t_{ter}$ & -100\\
    positive reward  $t_{move}$ & 1\\
    \hline
    \end{tabular}
\end{table}

Compared with the step-decreasing learning rate in the training of the supervised learning model \cite{tai2016deep}, here we use a much smaller fixed learning rate in the end-to-end training for the deep reinforcement learning model. As the only feedback to motivate the network convergence, the negative reward for the collision between the robot and obstacles must be very large as in \cite{tl_rcar_2016}. The training parameters are shown in Table~\ref{tab:training_parameters} in detail. All models are trained and tested with Caffe \cite{jia2014caffe} on a single NVIDIA GeForce GTX 690.
\begin{table}[h]
    \centering
    \caption{Speeds of different moving commands}
    \label{tab:moving_commands}
    \begin{tabular}{l | c c c c c | c }
    \hline
    \hline
        &\multicolumn{5}{|c|}{ angular velocity (rad/s) }  & line velocity  \\
       & Left & H-Left & Straight & H-right & Right& (m/s) \\
    \hline
    Train & 1.4 & 0.7&0 & -0.7& -1.4 & 0.32\\   
    \hline
    Test & 1.2& 0.6& 0& -0.6&-1.2 & 0.25\\
    \hline
    \end{tabular}
\end{table}

 \begin{table*}[!hb]
    \centering
    \caption{The average counts of moving steps and the average moving distances in every start point}
    \label{tab:table_heatmapscore}
    \begin{tabular}{l | l | c c c c c c c c c c c c }
    \hline
    \hline
     Metric&Model  & 1 & 2 & 3 & 4 & 5 & 6 & 7 & 8 & 9 & 10 & 11 & 12 \\
   \hline
   \multirow{6}{*}{Count} & SL &$ 7.6 $ &$ 16.4 $ &$ 33.4 $ &$ 3.0 $ &$ 21.1 $ &$ 20.2 $ &$ 6.9 $ &$ 14.0 $ &$ 16.7 $ &$ 5.6 $ &$ 10.6 $ &$ 19.6 $\\
                        & RL  &$ 16.7 $ &$ \mathbf{102.0} $ &$ 4.0 $ &$ 19.1 $ &$ 14.5 $ &$ 7.4 $ &$ 5.0 $ &$ 3.0 $ &$ 16.8 $ &$ 7.0 $ &$ 18.4 $ &$ 36.6 $\\
    & DRL 500 &$ 6.6 $ &$ 3.7 $ &$ 5.8 $ &$ 4.0 $ &$ 5.0 $ &$ 3.5 $ &$ 4.0 $ &$ 9.8 $ &$ 8.3 $ &$ 4.0 $ &$ 21.0 $ &$ 36.3 $\\
    & DRL 4000 &$ 16.2 $ &$ 13.7 $ &$ 28.7 $ &$ 26.2 $ &$ 26.2 $ &$ \mathbf{26.9} $ &$ 10.4 $ &$ 41.8 $ &$ 13.0 $ &$ 20.3 $ &$ 12.4 $ &$ 23.1 $\\
    & DRL 7500 &$ 44.3 $ &$ 38.0 $ &$ 16.4 $ &$ 31.1 $ &$ 23.4 $ &$ 23.2 $ &$ \mathbf{29.1} $ &$ 30.8 $ &$ 18.1 $ &$ 24.6 $ &$ 27.7 $ &$ 21.4 $ \\
    & DRL 40000 &$ \mathbf{135.9} $ &$ 71.9 $ & $ \mathbf{66.1} $ & $ \mathbf{151.5} $ &$ \mathbf{91.7} $ &$ 17.8 $ &  $14.3 $ &$ \mathbf{102.7} $ &$ \mathbf{158.2} $ &$ \mathbf{86.0} $ &$ \mathbf{101.9} $ &$ \mathbf{113.0} $\\
 \hline    
   \multirow{6}{*}{Distance} & SL &$ 1.1 $ &$ 2.1 $ &$ 7.3 $ &$ 0.2 $ &$ 3.8 $ &$ 6.1 $ &$ 0.8 $ &$ 2.6 $ &$ 2.6 $ &$ 0.5 $ &$ 1.6 $ &$ 3.2 $ \\
                        & RL &$ 3.6 $ &$ 3.0 $ &$ 0.2 $ &$ 5.4 $ &$ 1.8 $ &$ 0.7 $ &$ 0.2 $ &$ 0.2 $ &$ 4.1 $ &$ 0.6 $ &$ 3.9 $ &$ 9.0 $\\
     & DRL 500 &$ 1.1 $ &$ 0.5 $ &$ 0.5 $ &$ 0.6 $ &$ 0.7 $ &$ 0.5 $ &$ 0.5 $ &$ 0.7 $ &$ 0.8 $ &$ 0.6 $ &$ 0.8 $ &$ 0.9 $\\     
    & DRL 4000 &$ 7.5 $ &$ 4.4 $ &$ \mathbf{17.6} $ &$ 10.3 $ &$ \mathbf{15.8} $ &$ \mathbf{10.7} $ &$ 2.1 $ &$ 20.3 $ &$ 2.4 $ &$ 5.7 $ &$ 2.8 $ &$ 10.1 $\\                 
    & DRL 7500 &$ 11.0 $ &$ \mathbf{20.5} $ &$ 6.0 $ &$ 10.6 $ &$ 9.4 $ &$ 9.0 $ &$ \mathbf{15.3} $ &$ 12.8 $ &$ 4.1 $ &$ 8.3 $ &$ 5.5 $ &$ 7.4 $\\
    & DRL 40000 &$ \mathbf{39.5} $ &$ 18.6 $ &$ 14.3 $ &$ \mathbf{21.0} $ &$ 7.6 $ &$ 10.5 $ &$ 2.7 $ &$ \mathbf{50.2} $ &$ \mathbf{33.0} $ &$ \mathbf{67.1} $ &$ \mathbf{19.2} $ &$ \mathbf{39.9} $\\
    \hline
    \end{tabular}
\end{table*}
Table~\ref{tab:moving_commands} lists the assignments of speeds for the five output moving commands both in training and testing procedures. All of the training or testing commands have the same line velocity. The various moving directions are declared with different angular velocities. The speeds for training procedure are a little larger than the speeds for testing. With a higher training speed, the robot is motivated to collide aggressively and there would be more samples with negative rewards in the replay memory. In the testing procedure, a small speed can keep the robot to make decisions more frequently.

   \begin{figure}[!h]
      \centering
      \includegraphics[width=0.8\columnwidth]{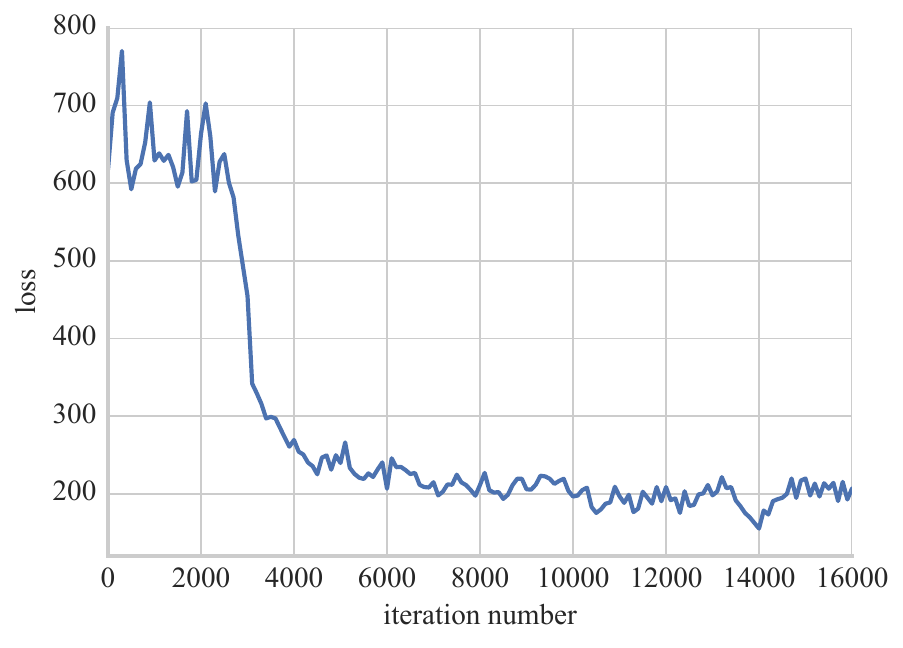}
      \caption{The loss decreasing curve as the training iterating. There is a batch of 32 samples used to do back-propagation in every iteration step.}
      \label{fig:training loss}
   \end{figure}

Fig.~\ref{fig:training loss} presents the loss reduction along training iterations. At each iteration step, a batch including 32 depth images is randomly chosen from the replay memory. Not like the training of conventional supervised learning methods, the loss of deep reinforcement learning may not converge to zero. It depends on the declaration of the negative reward extremely. Among the estimation Q-values for state-action pairs, the maximal represents the optimal action. The value itself can limited present the sum of the future gains \cite{mnih2015human}. Seen in the figure, the loss converges after 4000 iterations. Test results of several trained models after 4000 iterations are compared in Section \ref{sec:analysis}. 

\subsection{Analysis of exploration tests} \label{sec:analysis}


We firstly look at the obstacle avoidance capability of the trained model. The trained deep reinforcement learning (DRL) models after 500, 4000, 7500, and 40000 iterations are chosen to test in the simulated environment. The trained supervised learning (SL) model from \cite{tai2016deep} and the reinforcement learning (RL) model from \cite{tl_rcar_2016} are compared directly without any revising for the model structure or tuning for the weights. In all the 12 start points shown in Fig.~\ref{fig:environment_figure}, every model starts 10 exploration episodes with the test speeds listed in Table~\ref{tab:moving_commands} for five moving commands. Besides, every test episode will stop automatically after 200 moving steps, so that the robot will not explore freely forever. With the same CNN structure for all trained models, the forward prediction takes $48(\pm5)ms$ for each raw depth input.
After the forward calculation for the real-time depth image received, the robot chooses the moving command with the highest evaluation. The average counts of moving steps for each start point are listed in Table~\ref{tab:table_heatmapscore}. The more of the moving steps, the longer time the robot has been freely moving in the simulated environment without collisions.
\begin{figure*}[!ht]
    \centering
    \subfigure[Supervised Learning]{\includegraphics[width=0.6\columnwidth]{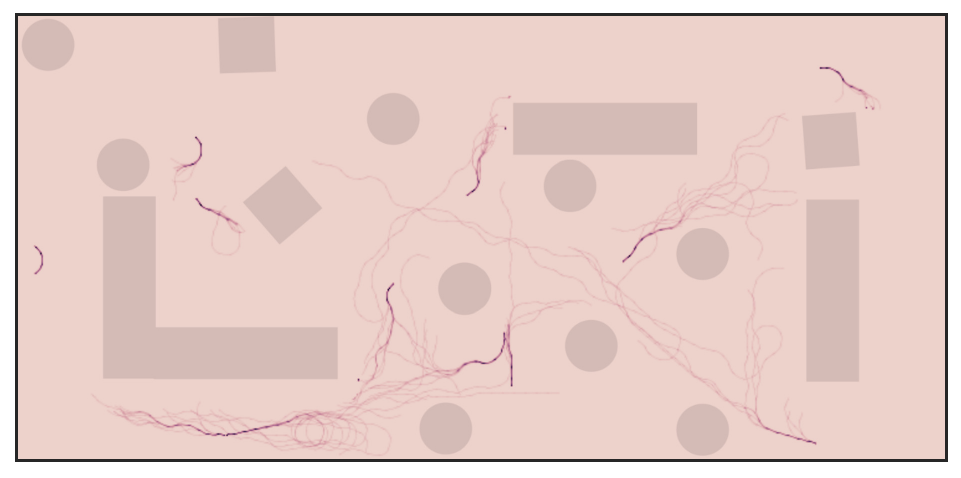}
    \label{fig:heatsl}}
    \subfigure[Reinforcement Learning]{\includegraphics[width=0.6\columnwidth]{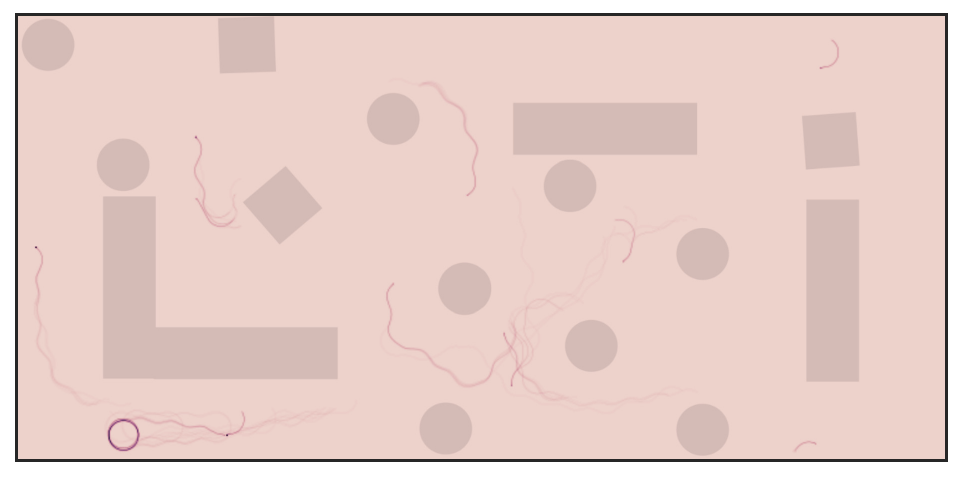}
    \label{fig:heatql}}
    \subfigure[DRL 500 iterations]{\includegraphics[width=0.6\columnwidth]{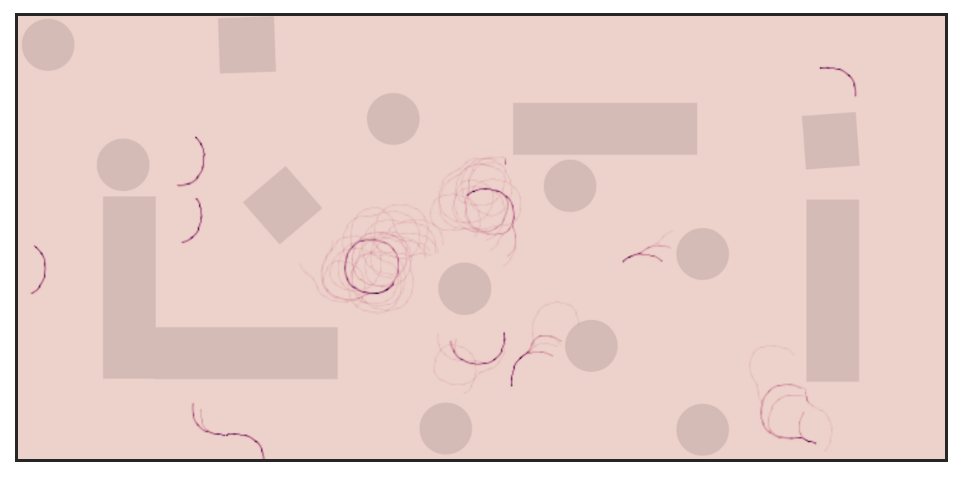}
    \label{fig:heat500}}
    \subfigure[DRL 4000 iterations]{\includegraphics[width=0.6\columnwidth]{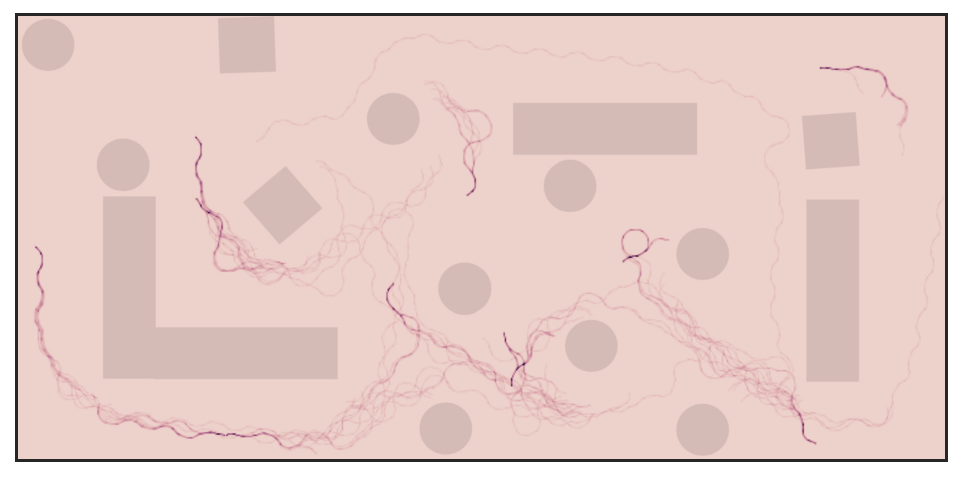}
    \label{fig:heat4000}}
    \subfigure[DRL 7500 iterations]{\includegraphics[width=0.6\columnwidth]{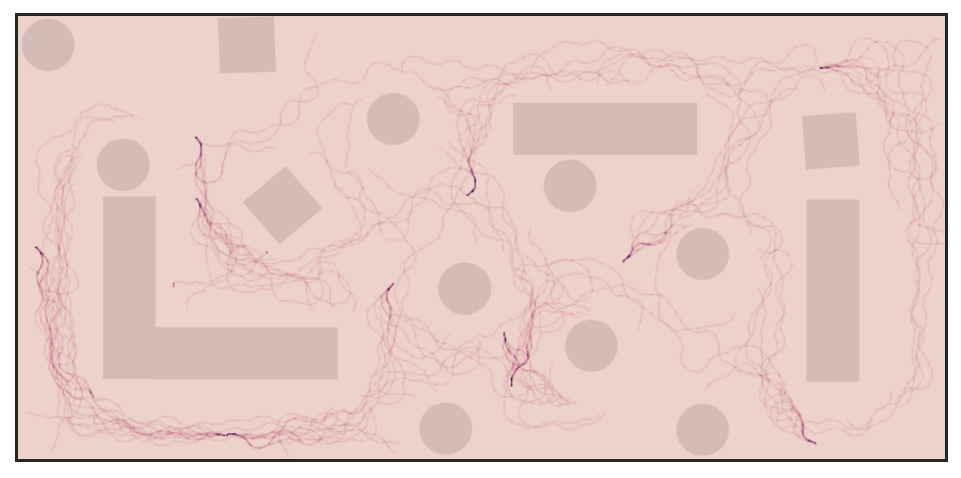}
    \label{fig:heat7500}}
    \subfigure[DRL 40000 iterations]{\includegraphics[width=0.6\columnwidth]{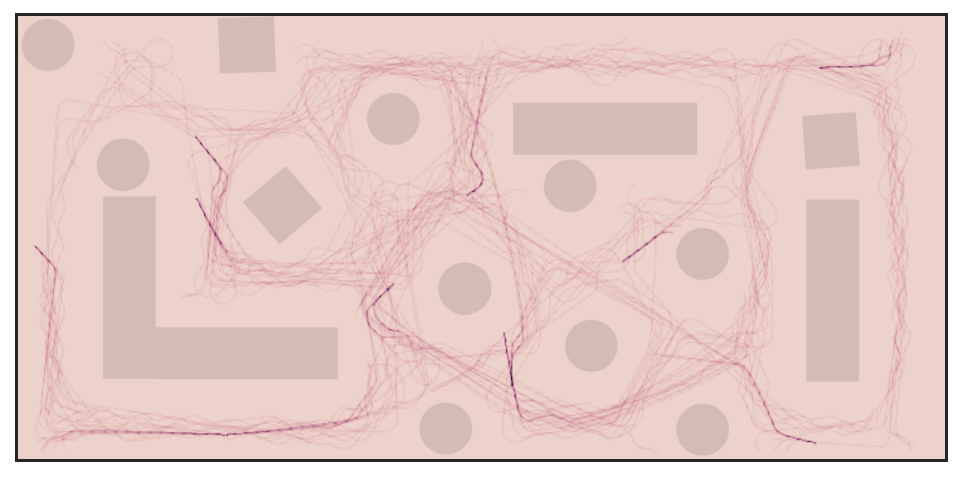}
    \label{fig:heat40000}}
    \caption{Heatmaps of the trajectory points' locations in the 10 test episodes of each model for all 12 start points. The counts of points in every map grid is normalized to [0,1]. Note that the circles at the left-bottom corner of (b) and the middle of (c) are actually a stack of circular trajectories caused by the actual motion of the robot.}
    \label{fig:heat_map}
\end{figure*}

Moving trajectory points of each model starting from all 12 positions are recorded. Fig.~\ref{fig:heat_map} depicts the trajectory after normalizing the counts of trajectory points to $[0,1]$ in each map grid of the training environment. The trained model may choose the \textit{left} or \textit{right} command to rotate in place so that there would not be any collisions happening, like the circles in the left-bottom corner of Fig.~\ref{fig:heatql} and the middle of Fig.~\ref{fig:heat500}. The very large moving count number of RL model in column 2 presented in Table~\ref{tab:table_heatmapscore} corresponds to this trajectory circle in Fig.~\ref{fig:heatql}. To avoid the appearance of this local minimal, the distances between the start and the end point are recorded as an additional evaluation metric. Notice that, after a long time exploration, the robot may move back to the area near the start point. So the distance may not be equal to the exploration ability of the trained model perfectly. 

From these heat maps, the supervised learning model cannot be adapted to the simulated environment especially in the scenes with multiple obstacles at different depths. The reinforcement learning model is the worst of the results. In the training of RL model \cite{tl_rcar_2016}, only the weights of the fully-connected network for policy iterations were updated iteratively. DRL model shows significant improvement compared with RL model because the training of DRL model is end-to-end. Thus, not only the policy network (\textit{fc} layers in Fig.~\ref{fig:network_structure}), but also the CNN model for feature representations is developed for complicated scenes.

Seen in Fig.~\ref{fig:heat_map}(c-f), the training for the exploration ability of DRL model is an online-learning process. In the 500-iteration case, the robot always chooses the same moving direction for any scenes. After 4000 iterations, it can be adapted to parts of the environment. In the 7500-iteration case, the robot can almost move freely in this whole simulated word. Furthermore, the robot usually chooses the optimal moving direction after 40000 iterations like the more efficient trajectories in Fig.~\ref{fig:heat40000}. Not like the fixed training datasets of RL model, newly collected training samples of DRL model are saved to replay memory increasingly. The evaluations of training scenes are calculated by the current model which is updated with the increasing of training iterations. Thus, the robot exploration ability will be increased over time.  

  
Comprehensively, the 40000-iteration case can almost explore in the trained environment completely. Considering the very long time training (12 hours) for 40000 iterations, we choose the model after 7500 iterations to analyze further. The training time for 7500 iterations is 2.5 hours. It confirms that the mobile robot can be adapted to an unfamiliar environment by transferring the weights of the pre-trained SL model to the DRL framework with a very short-time and end-to-end DRL training. 

\subsection{Analysis of receptive fields in the cognitive process}

Convolutional neural networks are usually considered to be black-box models. The internal activation mechanism of CNN is rarely analyzed. In \cite{zeiler2014visualizing}, the strongest activation areas of the feature representations are presented by backtracking the receptive field in the source input. We propose a backtracking method by multiplying the last layer of feature representations (\textit{pool3} in Fig.~\ref{fig:network_structure}) with a single channel convolutional filter.  The dimension of \textit{pool3} is $64 \times 20 \times 15 $ in this paper. Multiply it with a convolutional kernel sizing $1 \times 15 \times 15$, which is fixed with bilinear weights as the one used for upsampling of semantic segmentation in \cite{long2015fully}. After that, a $120 \times 160$ matrix is reproduced as the same size as the input image.

\begin{table}[!hb]
    \centering
    \caption{Evaluations of moving commands for different scenes}
    \label{tab:table_score}
    \begin{tabular}{ c| c | c c c c c}
    \hline
    \hline
      &  & Left & H-Left & Straight & H-Right & Right \\
    \hline
     \multirow{2}{*}{S1} & SL  &$ 23.9 $ &$ \mathbf{26.3} $ &$ -10.6 $ &$ -35.3 $ &$ -34.4 $ \\
            &DRL &$ \mathbf{-16.3} $ &$ -36.2 $ &$ -31.7 $ &$ -38.7 $ &$ -44.5 $ \\
    \hline     
     \multirow{2}{*}{S2} & SL  &$ -18.2 $ &$ \mathbf{-0.1} $ &$ 46.2 $ &$ -30.1 $ &$ -8.8 $ \\
                 &DRL &$ -30.0 $ &$ -25.4 $ &$ -19.8 $ &$ -21.2 $ &$ \mathbf{-15.3} $\\
       \hline
    \multirow{2}{*}{S3} &  SL  &$ 4.2 $ &$ 5.9 $ &$ -21.3 $ &$ \mathbf{7.7} $ &$ -0.2 $ \\
                &DRL &$ \mathbf{-5.3} $ &$ -15.3 $ &$ -13.0 $ &$ -16.5 $ &$ -19.9 $ \\
       \hline
         \multirow{2}{*}{S4} & SL  &$ -4.0 $ &$ -5.5 $ &$ -15.2 $ &$ \mathbf{13.3} $ &$ -0.4 $ \\
                &DRL &$ -4.6 $ &$ -5.1 $ &$ \mathbf{-3.8} $ &$ -4.8 $ &$ -4.3 $ \\
     \hline
     \multirow{2}{*}{S5} & SL  &$ -17.4 $ &$ 12.3 $ &$ \mathbf{23.0} $ &$ -9.7 $ &$ -18.4 $ \\
            &DRL &$ \mathbf{-9.9} $ &$ -19.2 $ &$ -15.8 $ &$ -19.2 $ &$ -21.6 $ \\
    \hline
       \hline
     \multirow{2}{*}{R1} & SL  &$ -38.2 $ &$ 17.5 $ &$ \mathbf{27.4} $ &$ -15.6 $ &$ -1.9 $ \\
            &DRL &$ -84.1 $ &$ -89.3 $ &$ -71.8 $ &$ -78.8 $ &$ \mathbf{-63.2} $ \\
    \hline     
     \multirow{2}{*}{R2} & SL  &$ -11.6 $ &$ \mathbf{45.6} $ &$ -50.3 $ &$ 4.1 $ &$ -8.3 $ \\
        &DRL &$ -8.3 $ &$ -9.3 $ &$ \mathbf{-7.1} $ &$ -8.6 $ &$ -7.5 $ \\
    \hline
     \multirow{2}{*}{R3} & SL &$ -18.4 $ &$ 32.2 $ &$ \mathbf{37.2} $ &$ -23.3 $ &$ -41.4 $ \\
        &DRL &$ \mathbf{-87.7} $ &$ -113.3 $ &$ -90.9 $ &$ -103.6 $ &$ -96.3 $ \\
       \hline
         \multirow{2}{*}{R4} & SL &$ 4.2 $ &$ 1.1 $ &$ -7.8 $ &$ -10.1 $ &$ \mathbf{15.2} $ \\
        &DRL &$ \mathbf{-87.6} $ &$ -141.1 $ &$ -115.2 $ &$ -134.8 $ &$ -137.4 $ \\
       \hline
         \multirow{2}{*}{R5}& SL  &$ 0.4 $ &$ -6.2 $ &$ -48.8 $ &$ \mathbf{32.8} $ &$ 11.9 $ \\
        &DRL    &$ -3.4 $ &$ -3.3 $ &$ \mathbf{-2.5} $ &$ -3.0 $ &$ -2.7 $ \\
    \hline
    \end{tabular}
\end{table}

\begin{figure*}[!ht]
    \centering
    \subfigure[Simulation environmental samples]{\includegraphics[width=\columnwidth]{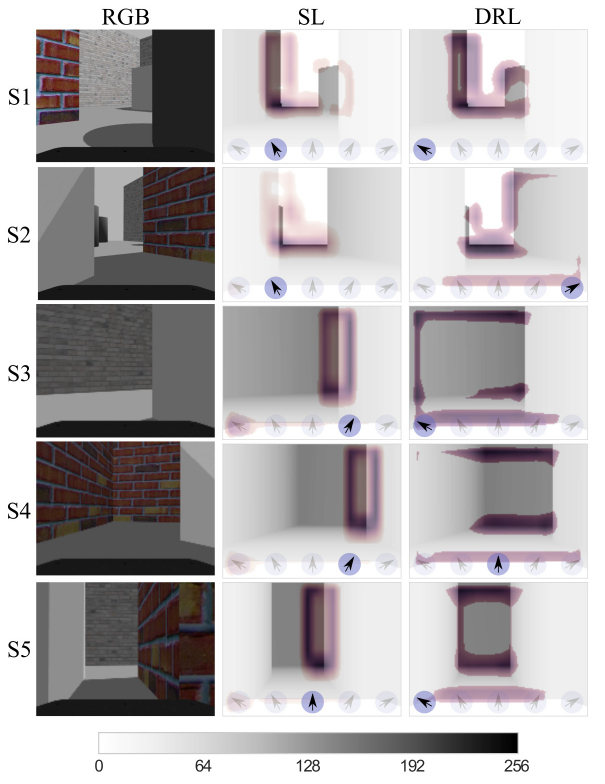}
    \label{fig:recep_simu}}
    \subfigure[Real world samples]{\includegraphics[width=\columnwidth]{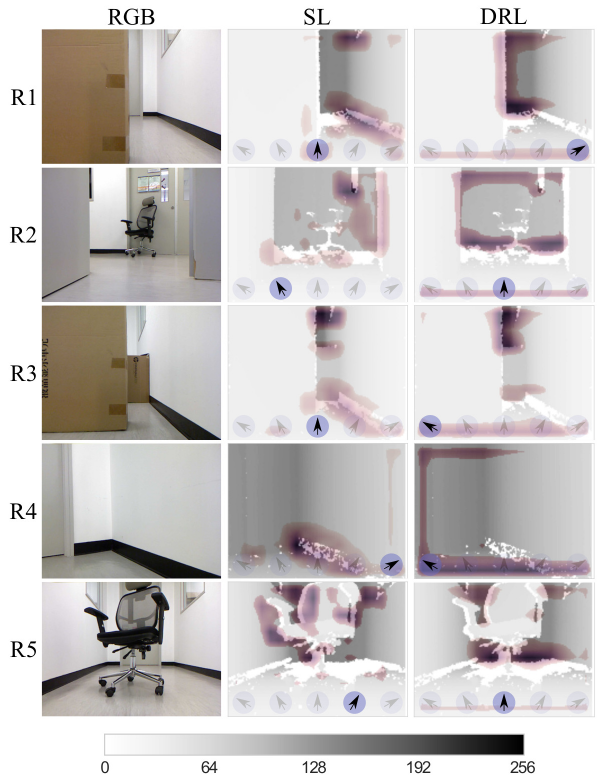}
    \label{fig:recep_real}}
    \caption{The receptive fields of the feature representations extracted by convolutional neural networks in both simulated environmental samples and real world samples. The purple area marked on the raw depth image represents the highest $10\%$ activation values. Both of the supervised learning model (SL) and the 7500-iteration deep reinforcement learning model (DRL) are compared. The arrow at the bottom of each receptive image is the chosen moving command based on evaluations listed in Table.~\ref{tab:table_score}. The left column shows the RGB images taken from the same scenes for references.}
    \label{fig:recep}
    \end{figure*}

We focus on the strongest activation area of the receptive matrix. Fig.~\ref{fig:recep_simu} shows the highest $ 10\% $ values of this matrix marked on the related raw depth images in five specific simulated training samples. The receptive fields of 7500-iteration DRL model are compared with the ones extracted from the SL model \cite{tai2016deep}. As mentioned above, these two models consist of the same convolutional structures. We choose five specific samples located in the fallible area based on the trajectory heat map of SL model shown in Fig.~\ref{fig:heatsl}. Before transported to \textit{Softmax} layer, feature representations of supervised learning model were firstly transformed to five values related to the five commands in \cite{tai2016deep}. These values are listed with the action-evaluations estimated by the 7500-iteration DRL model in Table~\ref{tab:table_score}. For both of the models, the highest value responds to the optimal moving command.

Notice that the area beyond the detection range of the \textit{kinect} camera is labeled as zero in the raw depth images. From Fig.~\ref{fig:recep_simu} and Table~\ref{tab:table_score}, for the SL model, the moving command towards the deepest area in range receives the highest output value naturally. It obviously motivates the convolutional model to activate the further area of the scenes, especially the junction part with the white untracked fields. In \textit{S1}, the furthest reflection can help the robot avoid the close obstacles. But when there are multiple level obstacles like in \textit{S2} and \textit{S3}, simply choosing the furthest part as the moving direction leads to collisions with nearby obstacles. Except for the furthest part, the DRL model also perceives the width of the route both in the nearby area and the furthest area as the several horizontal cognitive stripes in the figure. That means the end-to-end deep reinforcement learning dramatically tunes the initial CNN weights from the SL mode. As the evaluations listed in Table~\ref{tab:table_score}, the DRL model not only helps the robot avoid the instant obstacles, but also improves the traversable detection ability like in \textit{S4} and \textit{S5}. When the route in \textit{S4} is not wide enough to pass through, the DRL model chooses the fully turning moving command. However, the SL model always chooses the furthest part. 

To prove the robustness of the trained model, five samples collected from real world environment by a \textit{kinect} camera mounted on a real \textit{turtlebot} are also tested as shown in Fig.~\ref{fig:recep_real}. The related command evaluations for SL model and the DRL model as mentioned above are listed in Table~\ref{tab:table_score} as well. Notice that, these real world scenes are not included in the training datasets for the SL model \cite{tai2016deep} either. The receptive fields of the SL model are still mainly focusing on the furthest area. Output values in Table~\ref{tab:table_score} present the limited exploration ability of the SL model for theses untrained samples. For the DRL model, even though only trained in the simulated environment, it keeps showing the ability to track the width of the route for real world samples. In \textit{R1} and \textit{R2}, the trained DRL model successfully detects the traversable direction. In \textit{R3}, it avoids the narrow space which is not enough to pass. In \textit{R4}, it chooses the optimal moving direction to fully turn left. However, in \textit{R5}, when suffering irregular nearby obstacles which are not implemented in the training environment, it keeps tracking the width of the furthest area and failed to avoid this irregular obstacle. 

Another fact for real world tests is that the estimation of the action-value can reflect the future expectation to some extends. Estimation values of \textit{R3} and \textit{R4} listed in Table.~\ref{tab:table_score} for all moving commands are obviously less than values of other scenes. It corresponds to the higher probability of collision when there are nearby obstacles. 
\section{Conclusion} \label{sec:conclusion}
In this paper, the utility of the deep reinforcement learning framework for robot exploration is proved under end-to-end training. 
The framework comprises two parts, convolutional neural networks for feature representations and fully-connected networks for decision making. We initialized the weights of convolutional networks by a previously trained model based on our previous work \cite{tai2016deep}. The deep reinforcement learning model extends the cognitive ability of mobile robots for more complicated indoor environments in an efficient online-learning process continuously.  
Analysis of receptive fields indicates the crucial promotion of end-to-end deep reinforcement learning: feature representations extracted by convolutional networks are motivated substantially for the traversability of the mobile robots both in simulated and real environments.

There are many aspects to be developed in the future like building a more complicated unstructured environment. To navigate in more complicated even outdoor environments, the continues raw RGB images should also be considered as inputs like in \cite{mnih2015human} but not single depth image. The state-of-the-art CNN structures for RGB images like \textit{VGG} and \textit{ResNet} can substitute the CNN framework in our deep reinforcement learning model. The semantic extraction ability of CNN for RGB images has been fully proved \cite{long2015fully}. That may be very helpful for not only exploration but also the mapping capability of mobile robots. The revised deep reinforcement learning algorithms for continued control \cite{gu2016continuous} should also be considered to improve the learning efficiency. 


\bibliographystyle{IEEEtran}
\bibliography{DRL_bib}
\end{document}